\pdfoutput=1

\documentclass[11pt]{article}

\usepackage[]{EACL2023}

\usepackage{times}
\usepackage{latexsym}

\usepackage[T1]{fontenc}

\usepackage[utf8]{inputenc}

\usepackage{microtype}

\usepackage{inconsolata}

\usepackage{amsmath}

\usepackage{booktabs}

\usepackage{multirow}

\usepackage{diagbox}

\usepackage{stfloats} 

\usepackage{color}

\usepackage{colortbl}

\usepackage{amssymb}

\usepackage{graphicx} 

\usepackage{epstopdf}

\usepackage{amsthm}
\usepackage{natbib}
\title{Mining Effective Features Using Quantum Entropy for Humor Recognition}

\author{Yang Liu \and Yuexian Hou \\
College of Intelligence and Computing, Tianjin University, Tianjin, China \\
\texttt{\{lauyon,yxhou\}@tju.edu.cn} \\}

\begin{document}
    \maketitle
    \begin{abstract}
        Humor recognition has been studied with several different methods in the past years.
        However, existing studies on humor recognition do not understand the mechanisms that generate humor.
        In this paper, inspired by the incongruity theory, any joke can be divided into two components (the setup and the punchline).
        Both components have multiple possible semantics, and there is an incongruous relationship between them.
        We use density matrices to represent the semantic uncertainty of the setup and the punchline, respectively, and design Quantum Entropy Uncertainty (QE-Uncertainty) and Quantum Entropy Incongruity (QE-Incongruity) with the help of quantum entropy as features for humor recognition.
        The experimental results on the SemEval2021 Task 7 dataset show that the proposed features are more effective than the baselines for recognizing humorous and non-humorous texts.

    \end{abstract}

    \section{Introduction}
    Humor is one of the most distinctive features of human behavior and a sign of mental maturity \cite{pasquali1990learning}.
    The study of humor has received extensive attention in the fields of linguistics, philosophy, psychology, and sociology \cite{mihalcea2010computational}.
    Computational humor is of particular interest, with the potential to transform computers into creative and motivational tools \cite{Nijholt2003HumorMI}.

\begin{figure}[tbp]
        \centering
        \includegraphics[width=\columnwidth]{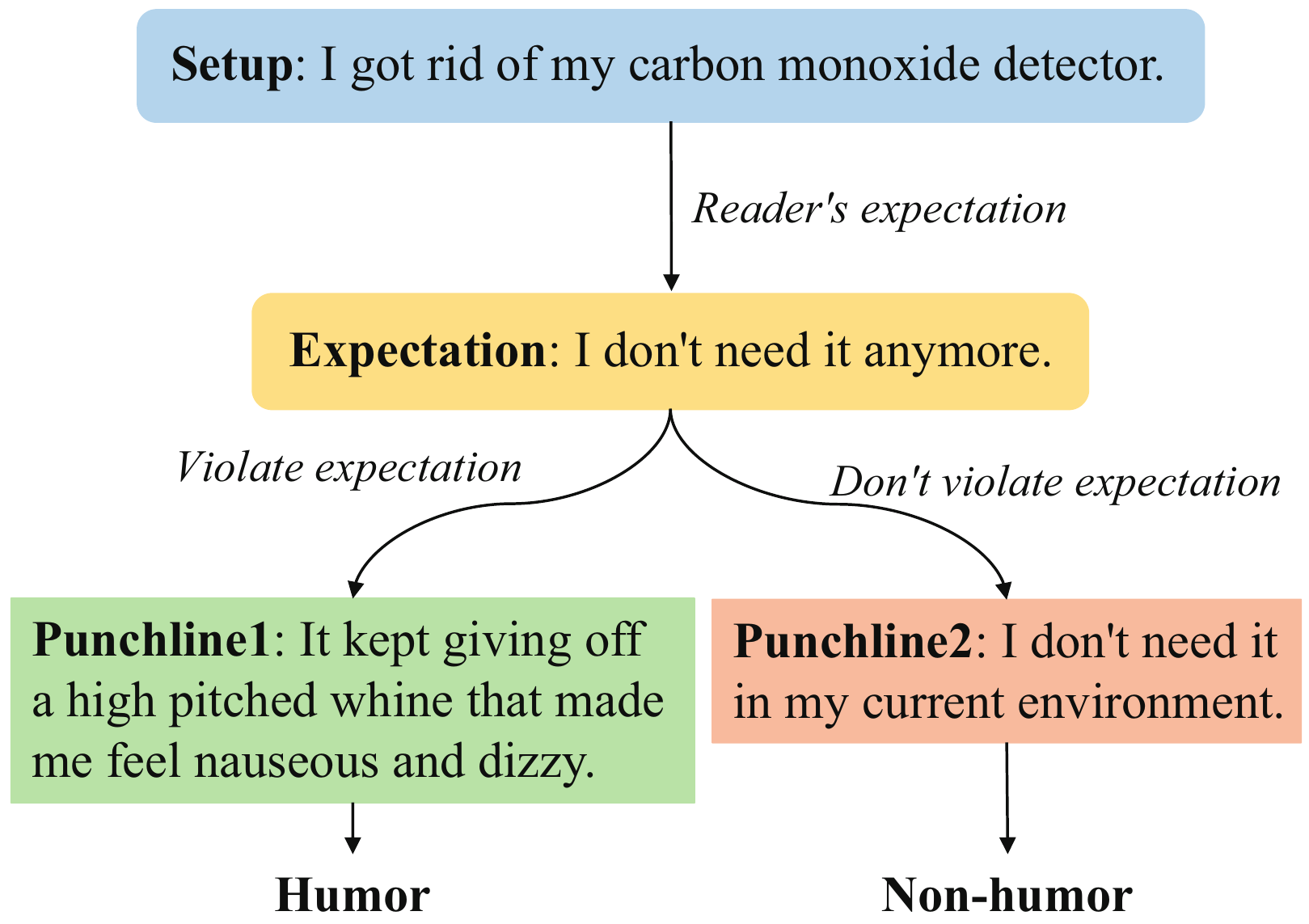}
        \caption{A humor and non-humor example containing the setup and the punchline.}
        \label{fig:humor_demo}
    \end{figure}

    This paper restricts research to humor recognition in computational humor, which aims to recognize whether a piece of text is humorous.
    As shown in Figure \ref{fig:humor_demo}, a joke usually includes two components: the setup and the punchline.
    The reader generates an expectation of the following text (the punchline) based on the content of the setup, and if the following text violates the reader's expectation, humor is generated, and vice versa.
    In fact, the incongruity theory of humor can explain the above process of producing humor.
    The incongruity theory states that humor is generated because a thing (the setup) has multiple underlying concepts, and there is an incongruity between the concept involved in the situation and the real object it represents (the punchline).

    Features based on semantic similarity \cite{mihalcea2010computational,yang-etal-2015-humor} and word association \cite{liu-etal-2018-modeling,cattle-ma-2018-recognizing} have achieved certain results, but they lack consideration of humorous mechanisms.
    \citet{xie-etal-2021-uncertainty} calculated the uncertainty and the surprisal values of the joke with the help of the GPT-2.
    But they did not model the semantic incongruity between the setup and the punchline.
    While the above approaches are somewhat effective, the incongruity theory requires us to model semantic uncertainty and the incongruity between the setup and the punchline.
    We take inspiration from quantum theory and use density matrices to represent the uncertainty of text semantics.
    Specifically, the setup and the punchline are represented as density matrices, respectively.
    Then, take the quantum entropy of the setup as \textbf{Q}uantum \textbf{E}ntropy \textbf{Uncertainty} (\textbf{QE-Uncertainty}) and the conditional quantum entropy between the setup and the punchline as \textbf{Q}uantum \textbf{E}ntropy \textbf{Incongruity} (\textbf{QE-Incongruity}).
    Experiments conducted on a manually-labeled dataset demonstrate that these two features are better than existing baselines in distinguishing between humorous and non-humorous texts, confirming the necessity of correlating semantic uncertainty with quantum theory.

    \section{Background}


    \subsection{The Incongruity Theory}

    The most widely accepted theory for explaining humor is the incongruity theory.
    The theory suggests that laughter is caused by an incongruity between the understanding of the text and its actual meaning \cite{mulder2002humour}.
    Immanuel Kant describes humor as ``\textit{the sudden transformation of a strained expectation into nothing} \citep{laurie2017masculinity}.''
    \citet{schopenhauer1966world} also believed that perceived incongruity exists between a concept and the real object it represents.
    The incongruity theory has also been developed in the field of linguistics.
    The Semantic Script-based Theory of Humor (SSTH) proposed by \citet{raskin1979semantic} is a scripted expression of the incongruity theory.
    SSTH is our bridge to mathematically model the incongruity theory.
    SSTH requires humorous texts to meet the following conditions:
    (1) The text is compatible, fully or in part, with two different (semantic) scripts.
    (2) The two scripts with which the text is compatible are opposite.

    \subsection{Density Matrix}

    \label{sec:desity_matrix}
    The mathematical form of quantum mechanics represents the probability space as a vector space (i.e., the Hilbert space $\mathbb{H}^n$) \citep{von2018mathematical}.
    Researchers often use Dirac's notation to represent unit vectors in this space.
    For example, a unit vector $\vec{u}$ and its transpose $\vec{u}^T$ are represented as $|u\rangle$ and $\langle u |$, respectively.
    The inner product of two unit vectors $|u\rangle$ and $|v\rangle$ is written as $\langle u|v \rangle$.
    The projector onto the direction $|u\rangle$ is its own outer product $|u\rangle  \langle u|$.
    The rank of each projector is one and each projector represents a quantum fundamental event, often called a \textit{dyad}.
    The density matrix \citep{nielsen2010quantum} is a generalization of the classical probability distribution.
    A density matrix $\rho$ can be defined as a mixture of dyads:
    \begin{equation}
        \rho = \displaystyle\sum_{i=1}^{n} p_i | \psi_i \rangle \langle \psi_i |
    \end{equation}
    where $|\psi_i \rangle$ represents a pure state with probability $p_i$.
    The density matrix $\rho$ is symmetric, positive semi-definite, and its trace is one.

    \subsection{Quantum Entropy}

    Quantum entropy is a generalization of the quantum case of classical Shannon entropy \citep{shannon1948mathematical}.
    If a quantum system is described by a density matrix $\rho$, its quantum entropy \cite{von2018mathematical} is defined as:
    \begin{equation}
        \mathrm{S}(\rho)=-\mathrm{tr}(\rho \ln \rho)
    \end{equation}

    The conditional quantum entropy \citep{PhysRevA.60.893} of the density matrix $\sigma$ given the known density matrix $\rho$ is defined as:
    \begin{equation}
        \begin{aligned}
            \mathrm{S}(\sigma | \rho) &=\mathrm{S}(\sigma \rho) - \mathrm{S}(\rho) \\
            &=-\mathrm{tr}(\sigma\rho \ln(\sigma\rho)) + \mathrm{tr}(\rho \ln \rho)
        \end{aligned}
    \end{equation}
    unlike classical conditional entropy, conditional quantum entropy can be negative.

    \section{Methodology}\label{sec:our_proposed_features}

    The incongruity theory holds that the prerequisite for humor is that the text has multiple semantic aspects.
    The reader does not understand one meaning but expects one while the punchline provides another, leading to incongruity.
    According to the incongruity theory, we should design features to represent the multiple semantic overlaps of the setup, as well as the incongruity of the semantics of the setup and the punchline.

    Normalize each word $w_i\in V$ as follows:
    \begin{equation}
        | w_i \rangle = \frac{\vec{w_i}}{ \lVert  \vec{w_i} \rVert  }
    \end{equation}
    where $\lVert \cdot \rVert$ represents the $L_2$-norm.
    The representation of each word can be viewed as a superposition in Hilbert space.

    A sentence of length $l$ is represented by an $n$-by-$n$ density matrix $\rho$:
    \begin{equation}
        \rho = \frac{1}{|l|}\sum_{i=1}^{l} | w_i \rangle \langle w_i |
    \end{equation}
    where the diagonal values of $\rho$ reflect the superposition semantics of sentences, and the non-diagonal values encode the correlation between semantics in a quantum way.

    \subsection{QE-Uncertainty}
    \label{sec:uncertainty}
    We take evidence of humor recognition from the setup, model the setup as a density matrix to represent its uncertainty semantics, and use the quantum entropy of the density matrix to represent the value of uncertainty.
    Formally, the QE-Uncertainty is calculated as follows:
    \begin{equation}
        \mathrm{U}(\rho)=-\mathrm{tr}(\rho \ln \rho)
    \end{equation}
    where $\rho$ represents the density matrix of the setup.
    The value of QE-Uncertainty reflects the amount of information contained in the text and the uncertainty of semantics.
    The larger the value, the more information the text contains, and the more likely the text is humorous.

    \subsection{QE-Incongruity}

    Another aspect of the incongruity theory is how different the semantics of the punchline is from expectations when the semantics of the setup are known (i.e., how much information we don't know about the punchline).
    In other words, how much information about the punchline is included in the setup?
    Specifically, the QE-Incongruity is defined as follows:
    \begin{equation}
        \begin{aligned}
            \mathrm{I}(\sigma | \rho) &= \mathrm{U}(\sigma\rho) - \mathrm{U}(\rho) \\
            &=- \mathrm{tr}(\sigma\rho \ln (\sigma\rho)) + \mathrm{tr}(\rho \ln \rho)
        \end{aligned}
    \end{equation}
    where $\rho$ and $\sigma$ represent the density matrices of the setup and the punchline, respectively.
    The value of QE-Incongruity describes how unknown the semantics of the punchline is when the setup is known.
    We argue that when the setup contains less semantics in the punchline, there will be incongruity, and there will be humor.

    \section{Related Work}

    The existing text humor recognition methods are mainly divided into feature-based methods and deep learning-based methods.
    \citet{mihalcea-strapparava-2005-making} use automatic classification techniques to integrate humor-specific features (alliteration, antonymy, slang) and content-based features into a machine-learning framework for humor classification tasks.
    \citet{mihalcea2010computational} divide the humor text into two components: the setup and the punchline.
    Humor recognition is performed by calculating the semantic correlation between the setup and the punchline based on the incongruity theory.
    \citet{morales-zhai-2017-identifying} use a generative language model combined with background text resources to construct multiple features to identify whether a comment is a humorous text.
    \citet{liu-etal-2018-modeling} combine discourse analysis and sentiment analysis to extract sentiment-related features to address humor recognition.
    \citet{xie-etal-2021-uncertainty} developed uncertainty and superisal with the help of the prediction results of the pre-trained language model GPT-2.
    In recent years, with the development of deep learning, some deep learning-based methods have been proposed.
    \citet{chen2017convolutional} use convolutional neural networks to identify humor in the TED talks corpus.
    \citet{chen-soo-2018-humor} used the highway network architecture to implement deep convolutional neural networks to predict humor on datasets of different types and different languages.
    \citet{weller-seppi-2019-humor} used pre-trained BERT for the humor classification task.
    \citet{fan2020phonetics}  combine the Bi-GRU network with phonetic structure and ambiguity for humor recognition.

    \section{Experiments}

    \subsection{Settings}

    We build a Support Vector Machine (SVM) classifier for humor classification.
    Experiments are performed on the SemEval 2021 Task 7\footnote{\href{https://semeval.github.io/SemEval2021/}{https://semeval.github.io/SemEval2021/}} dataset modified by \citet{xie-etal-2021-uncertainty}.
    The dataset consists of a total of 3,052 labeled samples, half of which are humor and the other half are non-humor.
    The text of each sample in the dataset is split into two parts (the setup and the punchline).
    For each sample in the dataset, the lengths of the setup and the punchline are both below 20, and the percentage of alphabetical letters is greater than 75\%, all of which start with alphabetical letters.
    We use Accuracy(Acc), Precision(P), Recall(R) and F1-Score(F1) as the evaluation metrics.
    P, R and F1 are macro-averaged.
    The experiments adopt 10-fold cross-validation, and the result is the average value of repeated experiments.

    \subsection{Baselines}

    Semantic similarity and semantic distance are the most commonly used text features, and we choose three such features as our baselines:

    \begin{itemize}
        \item \textbf{Path similarity} \cite{rada1989development} is a similarity measure based on the shortest path, defined as follows:
        \begin{equation}
            \mathrm{Sim}_{path} = \frac{ 1 }{ 1 + \mathrm{D}(c_1,c_2) }
        \end{equation}
        where $\mathrm{D}(c_1,c_2)$ represents the shortest path in WordNet between concepts $c_1$ and $c_2$.
        \item \textbf{Disconnection} \cite{yang-etal-2015-humor} is defined as the maximum distance between word pairs in the text.
        \item \textbf{Repetition} \cite{yang-etal-2015-humor} is defined as the minimum distance between word pairs in the text.
    \end{itemize}

    In addition, we consider two GPT-2 based features proposed by \citet{xie-etal-2021-uncertainty} as baselines.
    They feed the text into GPT-2 model to predict the next token. While predicting the tokens of $y$, GPT-2 produces a probability distribution $v_i$ over the vocabulary.

    \begin{itemize}
        \item \textbf{Uncertainty} is obtained by calculating the average entropy of the probability distribution $v_i$ on the vocabulary, defined as:
        \begin{equation}
            \mathrm{U}(x,y) = - \frac{1}{|y|} \displaystyle\sum_{i=1}^n\displaystyle\sum_{w \in V} v_i^w \log v_i^w
        \end{equation}
        where $n$ represents the length of $y$ and $V$ is the vocabulary.

        \item \textbf{Surprisal} describes the degree of surprise when the language model generates the punchline, which is defined as follows:
        \begin{equation}
            \begin{aligned}
                \mathrm{S}(x,y) &= - \frac{1}{|y|}\log \mathrm{p}(y|x)\\
                &=- \frac{1}{|y|} \displaystyle\sum_{i=1}^n \log v_i^{y_i}
            \end{aligned}
        \end{equation}
    \end{itemize}


    \subsection{Predict Using Individual Features}
    \label{sec:individual_feature_prediction}
    Table \ref{tab:individual_feature_prediction} shows the results of individual feature prediction.
    Compared with the baselines, our proposed features QE-Uncertainty and QE-Incongruity achieve higher scores on all four metrics, with QE-Incongruity achieving the best results.
    In particular, compared with Uncertainty based on classical Shannon entropy, QE-Uncertainty under our quantum framework is greatly improved.
    This shows the necessity of quantum generalization for semantic uncertainty problems.

    \begin{table}[!htbp]
        \small
        \centering
        \caption{Experimental results of individual features.
        The results for features with an asterisk are reported by \citet{xie-etal-2021-uncertainty}.}
        \label{tab:individual_feature_prediction}
        \begin{tabular}{lcccc}
            \toprule 
            Features       & P               & R               & F1              & Acc             \\
            \midrule 
            Random         & 0.5000          & 0.5000          & 0.5000          & 0.5000          \\
            \midrule
            Sim$_{path}$   & 0.5123          & 0.5070          & 0.4555          & 0.5062          \\
            Disconnection  & 0.6475          & 0.5503          & 0.4610          & 0.5501          \\
            Repetition     & 0.5592          & 0.5577          & 0.5538          & 0.5567          \\
            Uncertainty*   & 0.5840          & 0.5738          & 0.5593          & 0.5741          \\
            Surprisal*     & 0.5617          & 0.5565          & 0.5455          & 0.5570          \\
            \midrule
            QE-Uncertainty & 0.6589          & 0.6318          & 0.6146          & 0.6314          \\
            QE-Incongruity & \textbf{0.6690} & \textbf{0.6450} & \textbf{0.6319} & \textbf{0.6451} \\
            \bottomrule 
        \end{tabular}
    \end{table}

    \subsection{Boost a Content-Based Classifier}
    To demonstrate the effectiveness of our proposed features combined with content-based classifiers.
    We use the 50-dimensional GloVe \citep{pennington-etal-2014-glove} embedding as the baseline.
    We encode the setup and the punchline as the average of their respective word embeddings, resulting in two vectors with dimensions 50.
    Concatenate these two vectors with our features to form a vector with dimension 101.
    Finally, put it into an SVM classifier for humor classification.
    The results are shown in Table \ref{tab:glove_with}, our features achieve higher improvements on content-based classifiers compared to baselines.

    \begin{table}[!htbp]
        \small
        \centering
        \caption{Experimental results of concatenating a content-based classifier.
        The results for features with an asterisk are reported by \citet{xie-etal-2021-uncertainty}. }
        \label{tab:glove_with}
        \begin{tabular}{lcccc}
            \toprule 
            Features        & P               & R               & F1              & Acc             \\
            \midrule 
            GloVe           & 0.8233          & 0.8232          & 0.8229          & 0.8234          \\
            \midrule
            GloVe+Sim$_{path}$ & 0.8246          & 0.8246          & 0.8233          & 0.8237          \\
            GloVe+Discon.       & 0.8262          & 0.8264          & 0.8258          & 0.8263          \\
            GloVe+Repeti.       & 0.8239          & 0.8241          & 0.8237          & 0.8240          \\
            GloVe+U*            & 0.8355          & 0.8359          & 0.8353          & 0.8359          \\
            GloVe+S*            & 0.8331          & 0.8326          & 0.8321          & 0.8326          \\
            \midrule
            GloVe+QE-U          & 0.8361          & 0.8363          & 0.8355          & 0.8359          \\
            GloVe+QE-I          & \textbf{0.8363}          & \textbf{0.8365}          & \textbf{0.8356}          & \textbf{0.8360}          \\
            \bottomrule 
        \end{tabular}
    \end{table}

    \subsection{Feature Visualization}
\begin{figure}[t]
        \centering
        \includegraphics[width=\columnwidth]{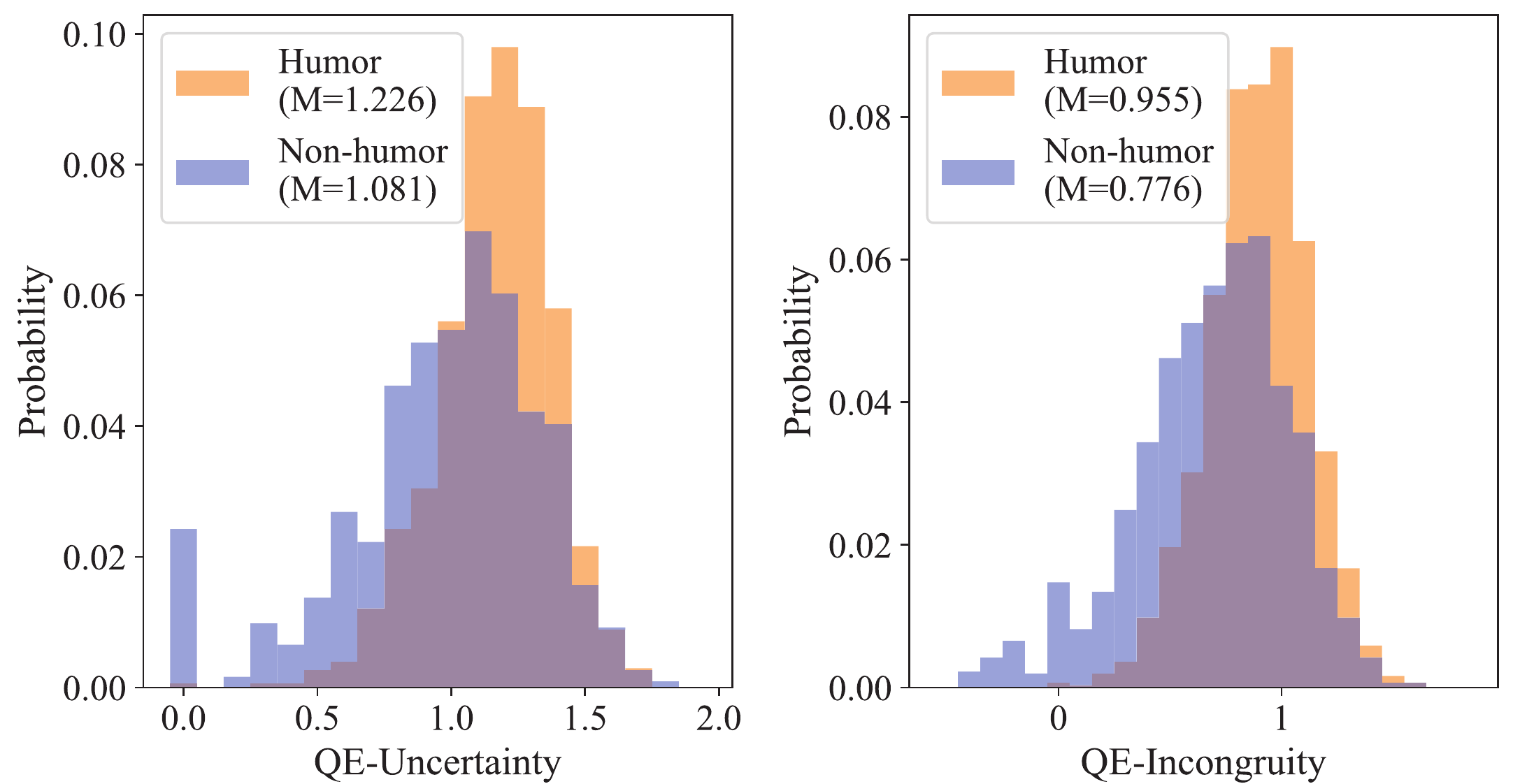}
        \caption{Histograms of our proposed features. The x-axis is the value of the feature, and the y-axis is the proportion of the feature in the total number of samples. \textbf{M} is the \textbf{M}edian of the current feature.}
        \label{fig:features_visualization}
    \end{figure}

    Figure \ref{fig:features_visualization} shows the distribution histograms of the values of QE-Uncertainty and QE-Incongruity for the joke and non-joke samples.
    From the figure, it can be found that jokes have higher QE-Uncertainty and QE-Incongruity values than non-jokes, which is consistent with what we stated in Section \ref{sec:our_proposed_features}.

    \section{Conclusion}
    In this paper, we model semantic uncertainty with a quantum framework.
    Inspired by the incongruity theory, we design two features, QE-Uncertainty and QE-Incongruity.
    We conduct experiments on the humor dataset, and the experimental results demonstrate the effectiveness of our proposed features.
    This suggests that the density matrix is an excellent framework for describing uncertainty and that the quantum entropy of the density matrix is a better feature to distinguish jokes from non-jokes than previously proposed features.
    We believe that the quantum framework can also be used for semantic uncertainty modeling for other tasks in the future.

    \section*{Limitations}
    In this paper, the density matrix representation of text is constructed in an averagely weighted manner, without considering the influence of weights on words.
    In addition, the density matrix as a text representation does not consider the position information of words.
    Furthermore, quantum generalization on the problem of multimodal humor recognition is also an interesting topic compared to unimodal humor recognition.

    \bibliography{anthology,custom}

\begin{thebibliography}{23}
\expandafter\ifx\csname natexlab\endcsname\relax\def\natexlab#1{#1}\fi

\bibitem[{Cattle and Ma(2018)}]{cattle-ma-2018-recognizing}
Andrew Cattle and Xiaojuan Ma. 2018.
\newblock \href {https://aclanthology.org/C18-1157} {Recognizing humour using
  word associations and humour anchor extraction}.
\newblock In \emph{Proceedings of the 27th International Conference on
  Computational Linguistics}, pages 1849--1858.

\bibitem[{Cerf and Adami(1999)}]{PhysRevA.60.893}
N.~J. Cerf and C.~Adami. 1999.
\newblock \href {https://doi.org/10.1103/PhysRevA.60.893} {Quantum extension of
  conditional probability}.
\newblock \emph{Phys. Rev. A}, 60:893--897.

\bibitem[{Chen and Lee(2017)}]{chen2017convolutional}
Lei Chen and Chong~Min Lee. 2017.
\newblock \href {https://doi.org/10.48550/arXiv.1702.02584} {Convolutional
  neural network for humor recognition}.
\newblock \emph{arXiv preprint arXiv:1702.02584v1}.

\bibitem[{Chen and Soo(2018)}]{chen-soo-2018-humor}
Peng-Yu Chen and Von-Wun Soo. 2018.
\newblock \href {https://aclanthology.org/N18-2018} {Humor recognition using
  deep learning}.
\newblock In \emph{Proceedings of the 2018 Conference of the North {A}merican
  Chapter of the Association for Computational Linguistics: Human Language
  Technologies, Volume 2 (Short Papers)}, pages 113--117.

\bibitem[{Fan et~al.(2020)Fan, Lin, Yang, Diao, Shen, Chu, and
  Zhang}]{fan2020phonetics}
Xiaochao Fan, Hongfei Lin, Liang Yang, Yufeng Diao, Chen Shen, Yonghe Chu, and
  Tongxuan Zhang. 2020.
\newblock \href {https://doi.org/10.1155/2020/2509018} {Phonetics and ambiguity
  comprehension gated attention network for humor recognition}.
\newblock \emph{Complexity}, 2020:1--9.

\bibitem[{Hickey-Moody and Laurie(2017)}]{laurie2017masculinity}
Anna Hickey-Moody and Timothy Laurie. 2017.
\newblock \href
  {https://www.researchgate.net/publication/313344206_Masculinity_and_Ridicule}
  {Masculinity and ridicule}.
\newblock In \emph{Gender: laughter}, pages 215--228. Macmillan Reference USA.

\bibitem[{Liu et~al.(2018)Liu, Zhang, and Song}]{liu-etal-2018-modeling}
Lizhen Liu, Donghai Zhang, and Wei Song. 2018.
\newblock \href {https://aclanthology.org/P18-2093} {Modeling sentiment
  association in discourse for humor recognition}.
\newblock In \emph{Proceedings of the 56th Annual Meeting of the Association
  for Computational Linguistics (Volume 2: Short Papers)}, pages 586--591.

\bibitem[{Mihalcea and Strapparava(2005)}]{mihalcea-strapparava-2005-making}
Rada Mihalcea and Carlo Strapparava. 2005.
\newblock \href {https://aclanthology.org/H05-1067} {Making computers laugh:
  Investigations in automatic humor recognition}.
\newblock In \emph{Proceedings of Human Language Technology Conference and
  Conference on Empirical Methods in Natural Language Processing}, pages
  531--538.

\bibitem[{Mihalcea et~al.(2010)Mihalcea, Strapparava, and
  Pulman}]{mihalcea2010computational}
Rada Mihalcea, Carlo Strapparava, and Stephen Pulman. 2010.
\newblock \href {https://doi.org/10.1007/978-3-642-12116-6_30} {Computational
  models for incongruity detection in humour}.
\newblock In \emph{Computational Linguistics and Intelligent Text Processing},
  pages 364--374.

\bibitem[{Morales and Zhai(2017)}]{morales-zhai-2017-identifying}
Alex Morales and Chengxiang Zhai. 2017.
\newblock \href {https://aclanthology.org/D17-1051} {Identifying humor in
  reviews using background text sources}.
\newblock In \emph{Proceedings of the 2017 Conference on Empirical Methods in
  Natural Language Processing}, pages 492--501.

\bibitem[{Mulder and Nijholt(2002)}]{mulder2002humour}
Mauk Mulder and Antinus Nijholt. 2002.
\newblock \href
  {https://wwwhome.ewi.utwente.nl/~anijholt/artikelen/ctit24_2002.pdf} {Humour
  research: State of the art}.
\newblock \emph{CTIT Technical Report Series}, (02-34):1--24.

\bibitem[{Nielsen and Chuang(2010)}]{nielsen2010quantum}
Michael~A Nielsen and Isaac~L Chuang. 2010.
\newblock \href {https://doi.org/10.1017/CBO9780511976667} {\emph{Quantum
  Computation and Quantum Information}}.
\newblock Cambridge University Press.

\bibitem[{Nijholt et~al.(2003)Nijholt, Stock, Dix, and
  Morkes}]{Nijholt2003HumorMI}
Anton Nijholt, Oliviero Stock, Alan Dix, and John Morkes. 2003.
\newblock \href {https://doi.org/10.1145/765891.766143} {Humor modeling in the
  interface}.
\newblock In \emph{CHI '03 Extended Abstracts on Human Factors in Computing
  Systems}, page 1050–1051.

\bibitem[{Pasquali(1990)}]{pasquali1990learning}
EA~Pasquali. 1990.
\newblock \href {https://doi.org/10.3928/0279-3695-19900301-10} {Learning to
  laugh: humor as therapy.}
\newblock \emph{Journal of Psychosocial Nursing and Mental Health Services},
  28(3):31--35.

\bibitem[{Pennington et~al.(2014)Pennington, Socher, and
  Manning}]{pennington-etal-2014-glove}
Jeffrey Pennington, Richard Socher, and Christopher Manning. 2014.
\newblock \href {https://aclanthology.org/D14-1162} {{G}lo{V}e: Global vectors
  for word representation}.
\newblock In \emph{Proceedings of the 2014 Conference on Empirical Methods in
  Natural Language Processing ({EMNLP})}, pages 1532--1543.

\bibitem[{Rada et~al.(1989)Rada, Mili, Bicknell, and
  Blettner}]{rada1989development}
Roy Rada, Hafedh Mili, Ellen Bicknell, and Maria Blettner. 1989.
\newblock \href {https://doi.org/10.1109/21.24528} {Development and application
  of a metric on semantic nets}.
\newblock \emph{IEEE transactions on systems, man, and cybernetics},
  19(1):17--30.

\bibitem[{Raskin(1979)}]{raskin1979semantic}
Victor Raskin. 1979.
\newblock \href {https://doi.org/10.3765/bls.v5i0.2164} {Semantic mechanisms of
  humor}.
\newblock In \emph{Annual Meeting of the Berkeley Linguistics Society}, pages
  325--335.

\bibitem[{Schopenhauer(1966)}]{schopenhauer1966world}
A.~Schopenhauer. 1966.
\newblock \href {https://books.google.com.hk/books?id=KM6XiJqLhucC} {\emph{The
  World as Will and Representation}}.
\newblock Dover books on philosophy. Dover Publications.

\bibitem[{Shannon(1948)}]{shannon1948mathematical}
Claude~Elwood Shannon. 1948.
\newblock \href {https://doi.org/10.1002/j.1538-7305.1948.tb01338.x} {A
  mathematical theory of communication}.
\newblock \emph{The Bell system technical journal}, 27(3):379--423.

\bibitem[{Von~Neumann(2018)}]{von2018mathematical}
John Von~Neumann. 2018.
\newblock \href
  {https://books.google.com.hk/books?hl=en&lr=&id=B3OYDwAAQBAJ&oi=fnd&pg=PR1&ots=tks1BC7DOR&sig=_jQL8aYPQxF05jJbVJtibSN1dcU&redir_esc=y#v=onepage&q&f=false}
  {\emph{Mathematical foundations of quantum mechanics: New edition}},
  volume~53.
\newblock Princeton university press.

\bibitem[{Weller and Seppi(2019)}]{weller-seppi-2019-humor}
Orion Weller and Kevin Seppi. 2019.
\newblock \href {https://aclanthology.org/D19-1372} {Humor detection: A
  transformer gets the last laugh}.
\newblock In \emph{Proceedings of the 2019 Conference on Empirical Methods in
  Natural Language Processing and the 9th International Joint Conference on
  Natural Language Processing (EMNLP-IJCNLP)}, pages 3621--3625.

\bibitem[{Xie et~al.(2021)Xie, Li, and Pu}]{xie-etal-2021-uncertainty}
Yubo Xie, Junze Li, and Pearl Pu. 2021.
\newblock \href {https://aclanthology.org/2021.acl-short.6} {Uncertainty and
  surprisal jointly deliver the punchline: Exploiting incongruity-based
  features for humor recognition}.
\newblock In \emph{Proceedings of the 59th Annual Meeting of the Association
  for Computational Linguistics and the 11th International Joint Conference on
  Natural Language Processing (Volume 2: Short Papers)}, pages 33--39.

\bibitem[{Yang et~al.(2015)Yang, Lavie, Dyer, and Hovy}]{yang-etal-2015-humor}
Diyi Yang, Alon Lavie, Chris Dyer, and Eduard Hovy. 2015.
\newblock \href {https://aclanthology.org/D15-1284} {Humor recognition and
  humor anchor extraction}.
\newblock In \emph{Proceedings of the 2015 Conference on Empirical Methods in
  Natural Language Processing}, pages 2367--2376.

\end{thebibliography}
    \bibliographystyle{acl_natbib}

    \appendix



\end{document}